\documentclass[11pt,a4paper]{article}
\usepackage[hyperref]{acl2021}
\usepackage{times}
\usepackage{latexsym}
\usepackage{todonotes}

\newcommand{\chisq}{$\chi^2$}

\usepackage{microtype}

\aclfinalcopy 

\title{More Than Words: Collocation Tokenization for Latent Dirichlet Allocation Models}

\author{Jin Cheevaprawatdomrong \\
  Chulalongkorn University \\
  \texttt{jin236248@gmail.com} \\\And
  Alexandra Schofield\\
  Harvey Mudd College \\
  \texttt{xanda@cs.hmc.edu} \\\And
   Attapol T. Rutherford\\
   Chulalongkorn University\\
  \texttt{attapol.t@chula.ac.th} \\}

\date{}

\begin{document}
\maketitle

\begin{abstract}
Traditionally, Latent Dirichlet Allocation (LDA) ingests words in a collection of documents to discover their latent topics using word-document co-occurrences. However, it is unclear how to achieve the best results for languages without marked word boundaries such as Chinese and Thai. Here, we explore the use of Pearson's chi-squared (\chisq) test, \textit{t}-statistics, and Word Pair Encoding (WPE) to produce tokens as input to the LDA model. The \chisq, \textit{t} and WPE tokenizers are trained on Wikipedia text to look for words that should be grouped together, such as compound nouns, proper nouns, and complex event verbs. We propose a new metric for measuring the clustering quality in settings where the vocabularies of the models differ. Based on this metric 
and other established metrics, we show that topics trained with merged tokens result in topic keys that are clearer, more coherent, and more effective at distinguishing topics than those unmerged models.

\end{abstract}

\section{Introduction}
Latent Dirichlet allocation (LDA) models \citep{blei2003latent} provide useful insights into themes and trends in a large text collection through the unsupervised inference of topics, or probability distributions over unigram word types in the corpus. In this model, a topic is often interpreted based on its highest-probability words, with documents expressed in terms of proportions of each topic. Unfortunately, the context in which these tokens arise can be obscured in the bag-of-words rendering of text as unigram counts in documents. For instance, a topic with high probabilities of both ``coffee'' and ``table'' is tempting to interpret as focusing on the furniture item ``coffee table'', but both words could be frequent in a discussion of cafes containing no coffee tables. This problem is amplified in languages without marked word boundaries, such as Chinese and Thai: tokenizers for these languages may split conceptual units into segments that, while functional as standalone words, do not express the concept of the original text. Meaningful interpretation of topics can be lost without careful recombination of these words.

In this paper, we evaluate three techniques to merge multiple adjacent words into conceptually-unified phrasal tokens prior to LDA model inference: Pearson Chi-square test, \textit{t}-statistic, and word pair encoding (WPE).
We apply merging strategies to different language families including Indo-European language (English), Kra-Dai language (Thai) and Sinitic language (Chinese). Inspired by silhouette coefficients, we also introduce a new method to assess the coherence of topics in a setting with variable vocabularies caused by different pre-processing treatments, which was not possible with previously proposed methods.
Using this new metric and existing topic model evaluations, we find that all three approaches to merging adjacent words can improve the likelihood, coherence, and topic distinctiveness of LDA models.

\section{Related Work}

Despite their popularity in analyzing large amounts of text data, LDA models are notoriously complex to evaluate.
One must evaluate both the statistical fit of a model and the human-registered thematic coherence of the words found to arise in the high-probability words, or keys, of a topic, which may not correlate \cite{chang_reading_tea_leaves}.
Analyses often combine evaluations of fit \cite{wallach_evaluation_methods} and automated approximations of human judgments of coherence \cite{bouma2009normalized,mimno2011optimizing} based on mutual information, even with the expectation these may only somewhat correlate with true human judgments \cite{lau2014machine}.
A limitation of these existing approaches, however, is that they expect the vocabulary and tokenization to remain constant between two models.
For our evaluation, we use a normalized log likelihood approach to capture fit while accounting for changes in vocabulary \cite{schofield-mimno-2016-comparing}.


Pre-processing steps can meaningfully alter the results of the LDA models even in languages with good tokenization heuristics such as English \cite{schofield-mimno-2016-comparing,schofield2017pulling}.
We believe that languages that do not have clear tokenization standards deserve investigation into what kind of processing is appropriate.
Many works recognize that LDA results can be improved when input are including phrases 
\citep{lindsey2012phrase, yu2013phrase, el2014scalable, wang2016ptr, bin2018chinese, li2018efficient}.
We consider it valuable to specifically assess approaches to determining these phrases.

\section{Collocations and Word Pair Encoding}
\emph{Collocations} consist of two or more words that can express conventional meaning.
Since collocations can convey information about multi-word entities, context, and word usage, we hypothesize that the introduction of multi-word tokens, which capture collocations as unigrams through concatenation, can help achieve more useful and coherent topic models.
For languages that do not have clear word boundaries, there is a possible additional benefit to multi-word tokens: it can be hard to intuit whether inferred word boundaries will have a large impact on the final results. Merging adjacent words into 'multi-word' tokens may help remedy the potential problem of a segmentation that is not optimal for topic modeling purposes.

Many methods are possible to select collocations from tokenized text, such as frequency, mean and variance, and statistical hypothesis testing. In this paper, we evaluate Pearson's chi-squared test (\chisq) and the \textit{t}-statistic for word co-occurrence, two hypothesis tests to determine if two words are collocated significantly more than would occur randomly. To implement these tests, we use the NLTK package
to compute \cite{nltk}. We impose a minimum frequency in the corpus for each selected bigram: otherwise, top bigrams from the \chisq~ test will contain only exceptional rare words,
as these are expected to co-occur so rarely that even a few co-occurrences can trigger significance.

Taking inspiration from byte-pair encoding, or BPE, we propose an alternative to obtain \emph{word-pair encoding} (WPE) tokens. 
To do this, we first tokenize a large corpus and then collect bigram counts for all bigrams found in the corpus. Second, we merge the most frequent bigram to form a new WPE token. This new bigram is then treated as a word in all occurrences. Next, we continue to repeat the counting and merging process with one extra word type.
Finally, we obtain a vocabulary list of both unigram and WPE tokens.

\section{Evaluation Metrics} \label{evaluationmetrics}

\textbf{Held-Out Likelihood.} When multi-word phrases are converted to individual tokens, the number of tokens in the document decreases while the size of the corpus vocabulary increases. It is therefore illogical to compare the likelihoods of the word-token model and WPE-token model directly. 

In order to normalize the scores between the two models that do not have the exact same vocabulary and tokens, we use the log-likelihood ratio between the LDA model likelihood and the null (unigram) likelihood for each model. 
In other words, we normalize the LDA model likelihood ($\mathcal{L}_{\textrm{model}}$) by dividing it with the unigram likelihood ($\mathcal{L}_{\textrm{unigram}}$) as introduced by  \citet{schofield-mimno-2016-comparing}:





\begin{equation}\label{ptll}
	\textrm{PTLL}_\textrm{norm} 
	= \frac{\log \mathcal{L}_{\textrm{model}} - \log \mathcal{L}_{\textrm{unigram}}}{N}
\end{equation}


\textbf{New Metric: Concatenation-based Embedding Silhouette (CBES)} 
Previous measures of topic coherence rely on statistics from the training data and assume that the vocabularies are identical for both models, which is not the case for our settings. To address this, we propose a new metric called a \emph{concatenation-based embedding silhouette} (CBES), which measures the coherence within the same topic and also the distinguishability of different topics in the LDA results.
CBES extends silhouette coefficients \citep{rousseeuw1987silhouettes}, a common clustering evaluation metric, by projecting tokens and multiword tokens into the same space and computing the silhouette coefficients in this vector space in the usual way.

A good topic should have all of their topic keys close to each other and away from other words that do not belong in the same topic. Silhouette coefficients computed in this vector space capture exactly this. It is critical that embeddings from the two models that we want to compare must be from the same embedding space. We achieve this by concatenating each document with the versions containing \chisq, \textit{t}, and WPE collocations before training the embeddings. We use the \texttt{gensim} \citep{rehurek_lrec} implementation of with the Continuous Bag-of-Word (CBOW) algorithm \citep{mikolov_word2vec} to obtain word embeddings. 








\section{Experiments} \label{modelanddata}

We test our methods on various corpora in English, Thai, and Chinese (Table \ref{doctok-table}). 
The English corpora are drawn from The New York Times \citep{sandhaus2008new}, the Yelp Dataset \footnote{www.yelp.com/dataset}, and United States State of the Union addresses (1790 to 2018) divided into paragraphs
\footnote{www.kaggle.com/rtatman/state-of-the-union-corpus-1989-2017}.
The Thai data come from the news articles in Prachathai
\footnote{github.com/PyThaiNLP/prachathai-67k}, the restaurant reviews from Wongnai \footnote{www.kaggle.com/c/wongnai-challenge-review-rating-prediction}, the BEST corpus \footnote{thailang.nectec.or.th/downloadcenter}, 
and the Thai National Corpus \citep{tnc}. 
The Chinese data come from three corpora: the news articles from Chinanews, restaurant reviews from Dianping,
\footnote{github.com/zhangxiangxiao/glyph}
and the movie reviews from Douban
\footnote{www.kaggle.com/utmhikari/doubanmovieshortcomments}.
Each corpus is separated into 75\% training documents and 25\% test documents. 

\begin{table}[]
	\footnotesize
	\centering
	\begin{tabular}{|l|p{.25in}|p{.3in}|p{.3in}|p{.3in}|p{.3in}|}
    \hline
              &                 &  &  \multicolumn{3}{c|}{\%merged} \\ \cline{4-6}
    Corpus     & Docs & Tokens    & \multicolumn{1}{|c|}{\chisq} & \multicolumn{1}{|c|}{\textit{t}} & \multicolumn{1}{|c|}{WPE} \\ \hline
    NYTimes    & \multicolumn{1}{|r|}{80K}     & \multicolumn{1}{|r|}{2M}   & \multicolumn{1}{|r|}{15.43}   & \multicolumn{1}{|r|}{15.99}& \multicolumn{1}{|r|}{16.55}    \\
    SOTU       & \multicolumn{1}{|r|}{21K}     & \multicolumn{1}{|r|}{1M}   & \multicolumn{1}{|r|}{12.21}   & \multicolumn{1}{|r|}{12.90}& \multicolumn{1}{|r|}{13.53}    \\
    Yelp      & \multicolumn{1}{|r|}{200K}    & \multicolumn{1}{|r|}{13M}  & \multicolumn{1}{|r|}{9.47}    & \multicolumn{1}{|r|}{10.33}& \multicolumn{1}{|r|}{12.16}    \\ \hline
    TNC        & \multicolumn{1}{|r|}{2K}      & \multicolumn{1}{|r|}{4M}   & \multicolumn{1}{|r|}{11.43}   & \multicolumn{1}{|r|}{11.50}& \multicolumn{1}{|r|}{8.40}      \\
    BEST       & \multicolumn{1}{|r|}{4K}      & \multicolumn{1}{|r|}{6M}   & \multicolumn{1}{|r|}{12.43}   & \multicolumn{1}{|r|}{12.50}& \multicolumn{1}{|r|}{9.53}     \\
    Wongnai    & \multicolumn{1}{|r|}{40K}     & \multicolumn{1}{|r|}{8M}   & \multicolumn{1}{|r|}{5.68}   & \multicolumn{1}{|r|}{5.71}& \multicolumn{1}{|r|}{4.48}     \\
    Prachathai & \multicolumn{1}{|r|}{68K}     & \multicolumn{1}{|r|}{119M} & \multicolumn{1}{|r|}{13.41}   & \multicolumn{1}{|r|}{13.45}& \multicolumn{1}{|r|}{10.36}    \\ \hline
    Chinanews  & \multicolumn{1}{|r|}{100K}    & \multicolumn{1}{|r|}{3M}   & \multicolumn{1}{|r|}{12.62}   & \multicolumn{1}{|r|}{14.35}& \multicolumn{1}{|r|}{10.62}    \\
    Dianping   & \multicolumn{1}{|r|}{100K}    & \multicolumn{1}{|r|}{4M}   & \multicolumn{1}{|r|}{3.22}    & \multicolumn{1}{|r|}{3.84}& \multicolumn{1}{|r|}{2.59}     \\
    Douban     & \multicolumn{1}{|r|}{200K}    & \multicolumn{1}{|r|}{2M}   & \multicolumn{1}{|r|}{4.61}    & \multicolumn{1}{|r|}{5.18}& \multicolumn{1}{|r|}{3.99}     \\ \hline
    \end{tabular}
	\caption{\label{doctok-table} A survey of corpora providing the number of documents and tokens, as well as the percentage of unigram tokens merged using each approach.}
\end{table}

We train the \chisq, \textit{t} and WPE-based tokenizers for each language on Wikipedia articles for that language. For Thai and Chinese, we use the entire Wikipedia database, but for English we use the filtered Wiki103 dataset \citep{merity2016pointer}.
English, Thai, and Chinese documents are tokenized with NLTK \citep{nltk}, Attacut \citep{attacut}, and Stanford Word Segmenter \citep{tseng2005conditional} respectively.
We follow the same pre-precessing steps for the training and the test documents: lemmatize and lowercase in English, and remove stopwords, symbols and digits for all languages. 
We limit the \chisq, \textit{t} and WPE approaches to 100,000 types.
Note that the top \chisq~collocations are full of specific names and rare words from Wikipedia because they appear together more than they would do randomly (Figure~\ref{top50-figure}). We use MALLET \citep{McCallumMALLET2002} with the default hyperparameters to train and evaluate topic models in both word and multi-word documents with 10, 50, 100 topics.
We run the experiment 10 times for each combination of corpus, type of model (word, Chi, \textit{t} or WPE) and number of topics to compute the means of the normalized held-out likelihood and CBES explained in section \ref{evaluationmetrics}.

\begin{figure}[]
	\footnotesize
	\centering
	\begin{tabular}{|p{7cm}|}
		\hline
		\textbf{\chisq:} debes jugar, euskaltel euskadi, taare zameen, chetro ketl, hetch hetchy, ngwat mahop, mullum malarum, pazz jop, phnom penh, eisernen kreuzes, sirimalle chettu, kasa vubu, moondram pirai, gjems onstad, lettow vorbeck, pather panchali, ioann zlatoust, kud wafter, poquita ropa, viribus unitis \\ \hline
		\textbf{\textit{t}:} united states, new york, world war, km h, take place, miles km, los angeles, united kingdom, first time, high school, tropical storm, new zealand, war ii, video game, mph km, h mph, north america, air force, two years, peak number \\ \hline
		\textbf{WPE:} unite state, new york, take place, first time, unite kingdom, follow year, world war ii, also know, next day, new york city, high school, los angeles, north america, even though, new zealand, follow day, become first, also use, year old, take part \\
		\hline
	\end{tabular}
	\caption{\label{top50-figure} Different collocation scoring methods result in different top 20 English collocations.}
\end{figure}

\begin{figure}[]
	\footnotesize
	\centering
	\begin{tabular}{|p{7cm}|}
		\hline
		\textbf{Word:} court federal judge charge case former trial say rule today state sentence supreme prison justice accuse order law file jury \\ \hline
		\textbf{\chisq:} today rule washington say judge state law court federal case legal \textbf{supreme\_court} may ban order abortion lawyers allow violate laws \\ \hline
		\textbf{\textit{t}:} washington today state say rule judge federal law case ban may court \textbf{supreme\_court} violate seek right settlement abortion july allow \\ \hline
		\textbf{WPE:} judge federal today charge court case trial washington former say lawyer lawyers accuse \textbf{supreme\_court} rule state hear order file \textbf{federal\_judge} \\
		\hline
	\end{tabular}
	\caption{\label{sotu-topic-key-compare-figure} Topic keys about judges in State of the Union}
\end{figure}

\begin{table*}[ht]
    \footnotesize
	\centering
	\setlength\tabcolsep{4pt} 
	\begin{tabular}{|l|r|r|r|r|r|r|r|r|r|r|r|r|}
    \hline
    \multicolumn{1}{|c|}{} & \multicolumn{4}{c|}{10 topics}                                                  & \multicolumn{4}{c|}{50 topics}                                                  & \multicolumn{4}{c|}{100 topics}                                                 \\ \cline{2-13} 
    \multicolumn{1}{|c|}{} & \multicolumn{1}{c|}{Word} & \multicolumn{1}{c|}{\chisq} & \multicolumn{1}{c|}{\textit{t}} & \multicolumn{1}{c|}{WPE} & \multicolumn{1}{c|}{Word} & \multicolumn{1}{c|}{\chisq} & \multicolumn{1}{c|}{\textit{t}} & \multicolumn{1}{c|}{WPE} & \multicolumn{1}{c|}{Word} & \multicolumn{1}{c|}{\chisq} & \multicolumn{1}{c|}{\textit{t}} & \multicolumn{1}{c|}{WPE} \\ \hline
    NYTimes    & .3781     & \textbf{.4263} & \textbf{.6223} & \textbf{.4460} & .5595     & \textbf{.6105}  & \textbf{.6237}  & \textbf{.6439}  & .6091      & \textbf{.6501}  & \textbf{.6651}  & \textbf{.6867}  \\
    SOTU       & .2711     & \textbf{.3032} & \textbf{.3148} & \textbf{.3273} & .3867     & \textbf{.4240}  & \textbf{.4375}  & \textbf{.4573}  & .4153      & \textbf{.4444}  & \textbf{.4584}  & \textbf{.4821}  \\
    Yelp      & .1717     & \textbf{.1974} & \textbf{.2028} & \textbf{.2149} & .2846     & \textbf{.3226}  & \textbf{.3303}  & \textbf{.3506}  & .3201      & \textbf{.3586}  & \textbf{.3672}  & \textbf{.3883}  \\ \hline
    TNC        & .7614     & .7512          & .7468          & .7578          & 1.0214    & \textbf{1.0459} & \textbf{1.0484} & \textbf{1.0405} & 1.0735     & \textbf{1.1027} & \textbf{1.1073} & \textbf{1.0972} \\
    BEST       & .7029     & .6742          & .6773          & .6969          & .9210     & \textbf{.9293}  & \textbf{.9306}  & \textbf{.9358}  & .9899      & \textbf{1.0053} & \textbf{1.0067} & \textbf{1.0097} \\
    Wongnai    & .2014     & \textbf{.2125} & \textbf{.2141} & \textbf{.2102} & .3191     & \textbf{.3379}  & \textbf{.3378}  & \textbf{.3338}  & .3492      & \textbf{.3663}  & \textbf{.3675}  & .3443           \\
    Prachathai & .4366     & \textbf{.4723} & \textbf{.4730} & \textbf{.4659} & .7139     & \textbf{.7761}  & \textbf{.7783}  & \textbf{.7599}  & .8036      & \textbf{.8736}  & \textbf{.8761}  & \textbf{.8565}  \\ \hline
    Chinanews  & .5161     & \textbf{.5444} & \textbf{.6560} & \textbf{.5492} & .8114     & \textbf{.8392}  & \textbf{.9548}  & \textbf{.8544}  & .9186      & \textbf{.9353}  & \textbf{.9758}  & \textbf{.9620}  \\
    Dianping   & .2571     & \textbf{.2629} & \textbf{.2649} & \textbf{.2617} & .4087     & \textbf{.4144}  & \textbf{.4179}  & \textbf{.4536}  & .4538      & \textbf{.4594}  & \textbf{.4631}  & \textbf{.4969}  \\
    Douban     & .2974     & \textbf{.3027} & \textbf{.3079} & \textbf{.3046} & .4136     & \textbf{.4139}  & \textbf{.4211}  & \textbf{.4168}  & .4464      & .4417           & \textbf{.4492}  & .4451   \\ \hline
    \end{tabular}
	\caption{\label{ll-table} Normalized unigram log-likelihood per token improvement in collocation models. }
\end{table*}

\begin{table*}
    \footnotesize
	\centering
	\setlength\tabcolsep{4pt} 
	\begin{tabular}{|l|r|r|r|r|r|r|r|r|r|r|r|r|}
		\hline
    \multicolumn{1}{|c|}{} & \multicolumn{4}{c|}{10 topics}                                                  & \multicolumn{4}{c|}{50 topics}                                                  & \multicolumn{4}{c|}{100 topics}                                                 \\ \cline{2-13} 
    \multicolumn{1}{|c|}{} & \multicolumn{1}{c|}{Word} & \multicolumn{1}{c|}{\chisq} & \multicolumn{1}{c|}{\textit{t}} & \multicolumn{1}{c|}{WPE} & \multicolumn{1}{c|}{Word} & \multicolumn{1}{c|}{\chisq} & \multicolumn{1}{c|}{\textit{t}} & \multicolumn{1}{c|}{WPE} & \multicolumn{1}{c|}{Word} & \multicolumn{1}{c|}{\chisq} & \multicolumn{1}{c|}{\textit{t}} & \multicolumn{1}{c|}{WPE} \\ \hline
    NYTimes    & .0111     & \textbf{.0293}  & \textbf{.0374}  & \textbf{.0484}  & -.0509    & \textbf{-.0451} & \textbf{-.0415} & \textbf{-.0305} & -.0820     & \textbf{-.0768} & \textbf{-.0748} & \textbf{-.0690} \\
    SOTU       & -.0052    & \textbf{.0014}  & \textbf{.0081}  & \textbf{.0131}  & -.0603    & \textbf{-.0589} & \textbf{-.0573} & \textbf{-.0541} & -.0802     & \textbf{-.0784} & \textbf{-.0773} & \textbf{-.0731} \\
    Yelp      & -.0617    & \textbf{-.0524} & \textbf{-.0463} & \textbf{-.0374} & -.1103    & \textbf{-.1028} & \textbf{-.0981} & \textbf{-.0912} & -.1291     & \textbf{-.1249} & \textbf{-.1211} & \textbf{-.1141} \\ \hline
    TNC        & -.0223    & \textbf{-.0125} & \textbf{-.0161} & \textbf{-.0196} & -.0935    & \textbf{-.0810} & \textbf{-.0837} & \textbf{-.0862} & -.1134     & \textbf{-.1031} & \textbf{-.1041} & \textbf{-.1072} \\
    BEST       & -.0323    & \textbf{-.0171} & \textbf{-.0145} & \textbf{-.0261} & -.0984    & \textbf{-.0863} & \textbf{-.0851} & \textbf{-.0831} & -.1143     & \textbf{-.0987} & \textbf{-.1019} & \textbf{-.0986} \\
    Wongnai    & -.0618    & -.0658          & -.0676          & -.0655          & -.1417    & \textbf{-.1406} & \textbf{-.1404} & \textbf{-.1406} & -.1631     & \textbf{-.1591} & \textbf{-.1602} & \textbf{-.1616} \\
    Prachathai & -.0153    & \textbf{.0094}  & \textbf{.0073}  & \textbf{.0025}  & -.0814    & \textbf{-.0632} & \textbf{-.0600} & \textbf{-.0692} & -.1124     & \textbf{-.0922} & \textbf{-.0900} & \textbf{-.0995} \\ \hline
    Chinanews  & -.0003    & \textbf{.0091}  & \textbf{.0207}  & \textbf{.0222}  & -.0536    & \textbf{-.0530} & \textbf{-.0459} & \textbf{-.0406} & -.0694     & \textbf{-.0679} & \textbf{-.0582} & \textbf{-.0545} \\
    Dianping   & -.0614    & \textbf{-.0539} & \textbf{-.0518} & \textbf{-.0516} & -.0977    & \textbf{-.0940} & \textbf{-.0908} & \textbf{-.0933} & -.1143     & \textbf{-.1124} & \textbf{-.1126} & \textbf{-.1137} \\
    Douban     & .0072     & \textbf{.0106}  & \textbf{.0208}  & \textbf{.0149}  & -.0763    & \textbf{-.0747} & \textbf{-.0734} & \textbf{-.0721} & -.0994     & -.1010          & -.1002          & \textbf{-.0991} \\ \hline
    \end{tabular}
	\caption{\label{sil-table} Silhouette improvement in collocation models }
\end{table*}

\section{Results and Discussion}
In general, corpora containing news have higher percentages of merged words, while those containing restaurant and movie reviews tend to see lower percentages (Table~\ref{doctok-table}). This could be because the news corpora are in a similar domain to that of the Wikipedia which we use to build the list of co-occurring words.
In \chisq, \textit{t} and WPE models, where the input contains multi-word tokens, the results usually improve over the word-token model. Most of the exceptions are from corpora containing restaurant and movie reviews, in which the percentages of merged words are lower.  

The normalized log-likelihood per token of the multi-word models is generally higher than the word model across languages and corpora (Table \ref{ll-table}). This means the multi-word models are better than their corresponding word models in reproducing the statistics of the held-out data.
We also see a general improvement in coherence in multi-word models (Table \ref{sil-table}).
Further, the higher CBES score indicates that topic-keys are more semantically coherent and topics are more distinct.
The topic keys from multi-word models form a coherent conceptual unit (Table~\ref{sotu-topic-key-compare-figure}). 
We can see that \textit{supreme\_court} in the multi-word models is more meaningful than \textit{supreme} or {\em court} in the word model. Similarly, meaning of \textit{federal\_judge} is more precise than just \textit{federal} and \textit{judge}.

If we compare by looking at the topic-keys of the word and multi-word models, we can come up with similar topics because we as a human who understands English and have general knowledge of the world can make the connection based on surrounding topic-keys that \textit{soviet} and \textit{union}, or \textit{super} and \textit{bowl}, or \textit{new} and \textit{york} are part of connected words even though they are not explicitly merged.
However, if we want to use these topic-keys as input to other tasks such as query search or neural network modeling, it is useful to feed the merged tokens to be explicit that the \textit{bowl} here doesn't refer to the deep dish used for food, and that the \textit{union} here doesn't refer to the worker association. 

\section{Conclusion}
In this work, we improve the quality of LDA models by better processing the input text before training the model.
We found that all three approaches to select candidate multi-word tokens---Pearson's chi-squared test, \textit{t}-statistics, and word-pair encoding---improve the results of trained topic models across numerous metrics. We also propose a new evaluation metric necessary for evaluating LDA topic models in scenarios where pre-processing changes the corpus vocabulary. As a future direction, we would like to explore other collocation measures with applications to other language families that are morphologically rich. 

\bibliography{wpelda}

\begin{thebibliography}{24}
\expandafter\ifx\csname natexlab\endcsname\relax\def\natexlab#1{#1}\fi

\bibitem[{Aroonmanakun(2007)}]{tnc}
Wirote Aroonmanakun. 2007.
\newblock Creating the thai national corpus.
\newblock \emph{MANUSYA: Journal of Humanities}, 10(3):4--17.

\bibitem[{Bin et~al.(2018)Bin, HE, HU, and Cheng}]{bin2018chinese}
GE~Bin, Chun-hui HE, Sheng-ze HU, and GUO Cheng. 2018.
\newblock Chinese news hot subtopic discovery and recommendation method based
  on key phrase and the lda model.
\newblock \emph{DEStech Transactions on Engineering and Technology Research},
  (ecar).

\bibitem[{Bird(2006)}]{nltk}
Steven Bird. 2006.
\newblock Nltk: the natural language toolkit.
\newblock In \emph{Proceedings of the COLING/ACL 2006 Interactive Presentation
  Sessions}, pages 69--72.

\bibitem[{Blei et~al.(2003)Blei, Ng, and Jordan}]{blei2003latent}
David~M Blei, Andrew~Y Ng, and Michael~I Jordan. 2003.
\newblock Latent dirichlet allocation.
\newblock \emph{Journal of machine Learning research}, 3(Jan):993--1022.

\bibitem[{Bouma(2009)}]{bouma2009normalized}
Gerlof Bouma. 2009.
\newblock Normalized (pointwise) mutual information in collocation extraction.
\newblock \emph{Proceedings of GSCL}, pages 31--40.

\bibitem[{Chang et~al.(2009)Chang, Gerrish, Wang, Boyd-graber, and
  Blei}]{chang_reading_tea_leaves}
Jonathan Chang, Sean Gerrish, Chong Wang, Jordan Boyd-graber, and David Blei.
  2009.
\newblock \href
  {https://proceedings.neurips.cc/paper/2009/file/f92586a25bb3145facd64ab20fd554ff-Paper.pdf}
  {Reading tea leaves: How humans interpret topic models}.
\newblock In \emph{Advances in Neural Information Processing Systems},
  volume~22. Curran Associates, Inc.

\bibitem[{Chormai et~al.(2020)Chormai, Prasertsom, Cheevaprawatdomrong, and
  Rutherford}]{attacut}
Pattarawat Chormai, Ponrawee Prasertsom, Jin Cheevaprawatdomrong, and Attapol
  Rutherford. 2020.
\newblock Syllable-based neural thai word segmentation.
\newblock In \emph{Proceedings of the 28th International Conference on
  Computational Linguistics}, pages 4619--4637.

\bibitem[{El-Kishky et~al.(2014)El-Kishky, Song, Wang, Voss, and
  Han}]{el2014scalable}
Ahmed El-Kishky, Yanglei Song, Chi Wang, Clare Voss, and Jiawei Han. 2014.
\newblock Scalable topical phrase mining from text corpora.
\newblock \emph{arXiv preprint arXiv:1406.6312}.

\bibitem[{Lau et~al.(2014)Lau, Newman, and Baldwin}]{lau2014machine}
Jey~Han Lau, David Newman, and Timothy Baldwin. 2014.
\newblock Machine reading tea leaves: Automatically evaluating topic coherence
  and topic model quality.
\newblock In \emph{Proceedings of the 14th Conference of the European Chapter
  of the Association for Computational Linguistics}, pages 530--539.

\bibitem[{Li et~al.(2018)Li, Yang, Zhou, Wang, Liu, and
  Zhang}]{li2018efficient}
Bing Li, Xiaochun Yang, Rui Zhou, Bin Wang, Chengfei Liu, and Yanchun Zhang.
  2018.
\newblock An efficient method for high quality and cohesive topical phrase
  mining.
\newblock \emph{IEEE Transactions on Knowledge and Data Engineering},
  31(1):120--137.

\bibitem[{Lindsey et~al.(2012)Lindsey, Headden, and
  Stipicevic}]{lindsey2012phrase}
Robert Lindsey, William Headden, and Michael Stipicevic. 2012.
\newblock A phrase-discovering topic model using hierarchical pitman-yor
  processes.
\newblock In \emph{Proceedings of the 2012 Joint Conference on Empirical
  Methods in Natural Language Processing and Computational Natural Language
  Learning}, pages 214--222.

\bibitem[{McCallum(2002)}]{McCallumMALLET2002}
Andrew~Kachites McCallum. 2002.
\newblock Mallet: A machine learning for language toolkit.
\newblock Http://mallet.cs.umass.edu.

\bibitem[{Merity et~al.(2016)Merity, Xiong, Bradbury, and
  Socher}]{merity2016pointer}
Stephen Merity, Caiming Xiong, James Bradbury, and Richard Socher. 2016.
\newblock \href {http://arxiv.org/abs/1609.07843} {Pointer sentinel mixture
  models}.

\bibitem[{Mikolov et~al.(2013)Mikolov, Chen, Corrado, and
  Dean}]{mikolov_word2vec}
Tomas Mikolov, Kai Chen, Greg Corrado, and Jeffrey Dean. 2013.
\newblock Efficient estimation of word representations in vector space.
\newblock \emph{arXiv preprint arXiv:1301.3781}.

\bibitem[{Mimno et~al.(2011)Mimno, Wallach, Talley, Leenders, and
  McCallum}]{mimno2011optimizing}
David Mimno, Hanna Wallach, Edmund Talley, Miriam Leenders, and Andrew
  McCallum. 2011.
\newblock Optimizing semantic coherence in topic models.
\newblock In \emph{Proceedings of the 2011 Conference on Empirical Methods in
  Natural Language Processing}, pages 262--272.

\bibitem[{{\v R}eh{\r u}{\v r}ek and Sojka(2010)}]{rehurek_lrec}
Radim {\v R}eh{\r u}{\v r}ek and Petr Sojka. 2010.
\newblock {Software Framework for Topic Modelling with Large Corpora}.
\newblock In \emph{{Proceedings of the LREC 2010 Workshop on New Challenges for
  NLP Frameworks}}, pages 45--50, Valletta, Malta. ELRA.

\bibitem[{Rousseeuw(1987)}]{rousseeuw1987silhouettes}
Peter~J Rousseeuw. 1987.
\newblock Silhouettes: a graphical aid to the interpretation and validation of
  cluster analysis.
\newblock \emph{Journal of computational and applied mathematics}, 20:53--65.

\bibitem[{Sandhaus(2008)}]{sandhaus2008new}
Evan Sandhaus. 2008.
\newblock The new york times annotated corpus.
\newblock \emph{Linguistic Data Consortium, Philadelphia}, 6(12):e26752.

\bibitem[{Schofield et~al.(2017)Schofield, Magnusson, and
  Mimno}]{schofield2017pulling}
Alexandra Schofield, M{\aa}ns Magnusson, and David Mimno. 2017.
\newblock Pulling out the stops: Rethinking stopword removal for topic models.
\newblock In \emph{Proceedings of the 15th Conference of the European Chapter
  of the Association for Computational Linguistics: Volume 2, Short Papers},
  pages 432--436.

\bibitem[{Schofield and Mimno(2016)}]{schofield-mimno-2016-comparing}
Alexandra Schofield and David Mimno. 2016.
\newblock Comparing apples to apple: The effects of stemmers on topic models.
\newblock \emph{Transactions of the Association for Computational Linguistics},
  4:287--300.

\bibitem[{Tseng et~al.(2005)Tseng, Chang, Andrew, Jurafsky, and
  Manning}]{tseng2005conditional}
Huihsin Tseng, Pi-Chuan Chang, Galen Andrew, Dan Jurafsky, and Christopher~D
  Manning. 2005.
\newblock A conditional random field word segmenter for sighan bakeoff 2005.
\newblock In \emph{Proceedings of the fourth SIGHAN workshop on Chinese
  language Processing}.

\bibitem[{Wallach et~al.(2009)Wallach, Murray, Salakhutdinov, and
  Mimno}]{wallach_evaluation_methods}
Hanna~M. Wallach, Iain Murray, Ruslan Salakhutdinov, and David Mimno. 2009.
\newblock \href {https://doi.org/10.1145/1553374.1553515} {Evaluation methods
  for topic models}.
\newblock In \emph{Proceedings of the 26th Annual International Conference on
  Machine Learning}, ICML '09, page 1105–1112, New York, NY, USA. Association
  for Computing Machinery.

\bibitem[{Wang et~al.(2016)Wang, Zhao, and Huang}]{wang2016ptr}
Minmei Wang, Bo~Zhao, and Yihua Huang. 2016.
\newblock Ptr: phrase-based topical ranking for automatic keyphrase extraction
  in scientific publications.
\newblock In \emph{International Conference on Neural Information Processing},
  pages 120--128. Springer.

\bibitem[{Yu et~al.(2013)Yu, Johnson, and Kavuluru}]{yu2013phrase}
Zhiguo Yu, Todd~R Johnson, and Ramakanth Kavuluru. 2013.
\newblock Phrase based topic modeling for semantic information processing in
  biomedicine.
\newblock In \emph{2013 12th International Conference on Machine Learning and
  Applications}, volume~1, pages 440--445. IEEE.

\end{thebibliography}
\bibliographystyle{acl_natbib}

\end{document}